\pdfoutput=1
\documentclass[11pt]{article}
\usepackage{times}

\usepackage[]{acl}
\usepackage{xcolor}

\usepackage{latexsym}
\usepackage[T1]{fontenc}
\usepackage[utf8]{inputenc}
\usepackage{microtype}
\usepackage{amsmath, amssymb, amsfonts, bm}
\usepackage{booktabs}
\usepackage{multicol, multirow}
\usepackage{graphicx}
\usepackage{arydshln}
\usepackage{subcaption, caption}
\usepackage{mathtools}

\title{
Towards Countering Essentialism through Social Bias Reasoning}

\author{Emily Allaway$^{1,2}$ \quad
        Nina Taneja$^1$ \quad
       Sarah-Jane Leslie$^3$ \quad
        Maarten Sap$^{2,4}$\\
        $^1$Columbia University, USA\\
        $^2$Allen Institute for Artificial Intelligence, USA\\
        $^3$Princeton University, USA\\
        $^4$Carnegie Mellon University\\
        \texttt{eallaway@cs.columbia.edu}
        }

\begin{document}
\maketitle
\begin{abstract}
Essentialist beliefs (i.e., believing that members of the same group are fundamentally alike) play a central role in social stereotypes and can lead to harm when left unchallenged. In our work, we conduct exploratory studies into the task of countering essentialist beliefs (e.g., \textit{``liberals are stupid''}). Drawing on prior work from psychology and NLP, we construct five types of counterstatements and conduct human studies on the effectiveness of these different strategies. Our studies also investigate the role in choosing a counterstatement of the level of explicitness with which an essentialist belief is conveyed. We find that statements that broaden the scope of a stereotype (e.g., to other groups, as in \textit{``conservatives can also be stupid''}) are the most popular countering strategy. We conclude with a discussion of challenges and open questions for future work in this area (e.g., improving factuality, studying community-specific variation) and we emphasize the importance of work at the intersection of NLP and psychology.
\end{abstract}

\section{Introduction}
\textit{Essentialism}, i.e., the belief that members of the same group are fundamentally alike, plays a crucial role in how prejudices and biases about social and demographic groups are formed and expressed \citep{leslie2014carving}.
For example, the statement ``\textit{I speak English, I don't speak libt*rd}'' implies the belief that all ``\textit{liberals are stupid}.''
If left unchallenged,
statements with such essentializing implications can cause harm by perpetuating and reifying stereotypical beliefs about social groups~\cite{greenwald1995implicit,steele2011whistling,prentice2007psychological,rhodes2012cultural,leshin2021does}.

\begin{figure}[t]
    \centering
    \includegraphics[width=\columnwidth]{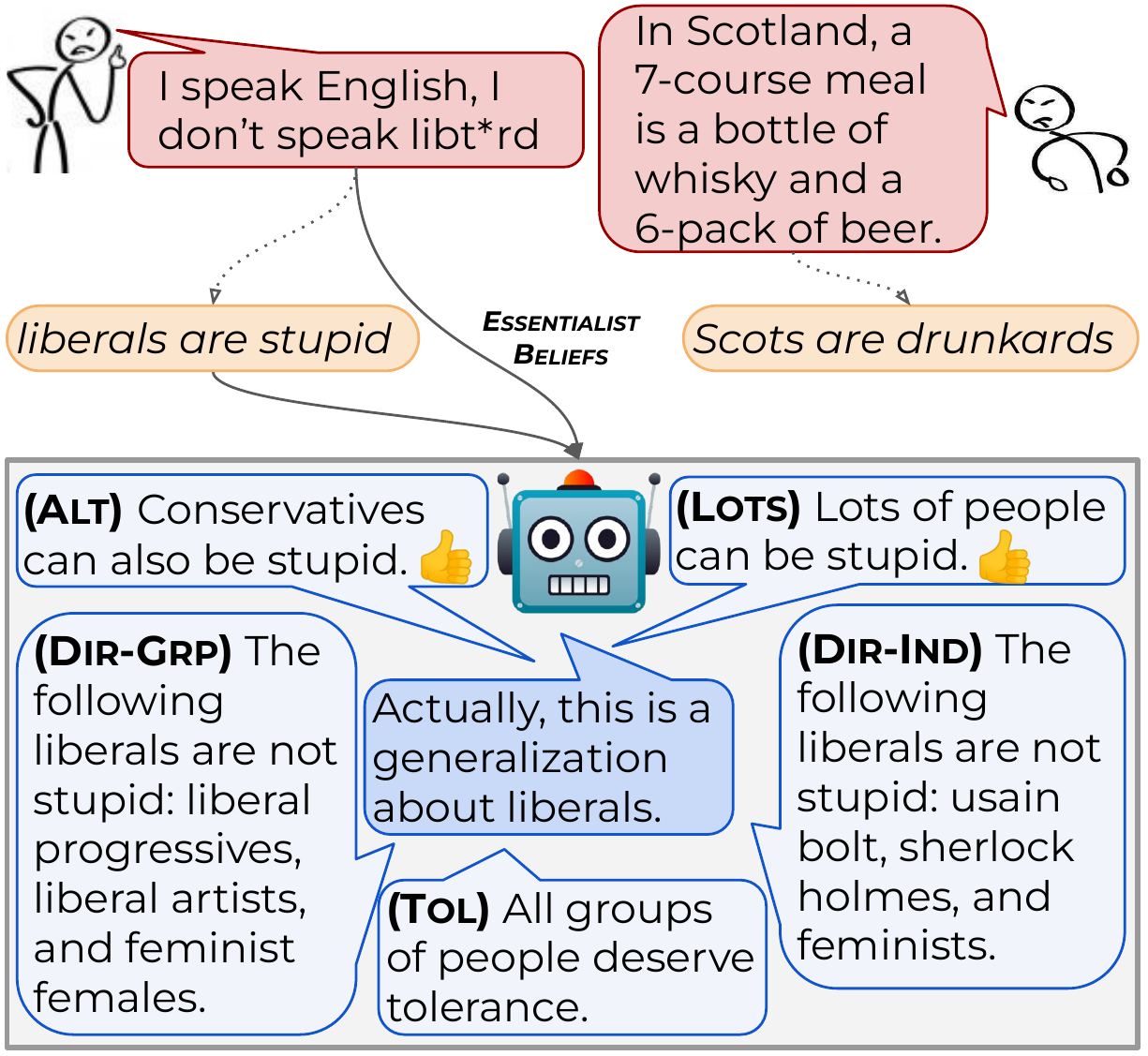}
    \caption{
    Two prejudiced statements with their essentialist implications (top), along with our five types of counterstatements automatically generated with our method.
    According to our results, a preferred strategy is highlighting that an implication applies to more than the targeted group (\textsc{Lots} or \textsc{Alts}).
    }
    \label{fig:intro}
\end{figure}

In this work, we investigate the task of combating essentialist statements and beliefs through psychologically and linguistically informed counterstatement generation. 
We examine these essentialist beliefs through the lens of \textit{generics}~\citep{rhodes2012cultural}, i.e., beliefs that attribute a quality to a target group without explicit quantification~\citep[``liberals are stupid'';][]{abelson1966subjective,carlson1995generic}.
In the context of toxic or hateful language, these generic beliefs can be both expressed directly or conveyed through subtle implications~\cite{gelman2003essential,sap2020socialbiasframes}.

Automatically countering essentialism is challenging because it requires deep psychological reasoning about the linguistic implications of statements -- for example, changing people's beliefs about stereotypes only through counterexamples is difficult~\cite{kunda1995maintaining}. Therefore, we examine five different strategies for combating essentializing stereotypes, combining insights from psychology~\cite{foster2016does,foster-hanson_leslie_rhodes_2019,wodak2015loaded} and NLP~\cite{allaway2022penguins}. We craft five types of statements (see Fig~\ref{fig:intro}): \textit{broadening} the scope of a stereotype by generalizing to ``all people'' or an alternative group (\textsc{Lots} and \textsc{Alt}), providing \textit{direct counter-evidence} through specific individuals or groups (\textsc{Dir-Ind} and \textsc{Dir-Grp}), and simply \textit{calling out} the generalization (\textsc{Tol}).
In contrast to prior studies on countering hate-speech which use uncontrolled end-to-end generation approaches ~\cite[e.g.,]{qian2019benchmark,Zhu2021GeneratePS,Chung2020ItalianCN}, we generate counterstatements by reasoning directly about the targeted group, attributed quality, and linguistic expression of a stereotype.

Since our work provides a preliminary exploration of this task, we conduct online studies in three settings where counterstatements are paired with human-written implications from the Social Bias Frames Inference Corpus (SBIC)~\cite{sap2020socialbiasframes}. In these settings, we explore variation in counterstatement effectiveness when the beliefs are conveyed either implicitly, explicitly without context, or as an explicit inference from provided context.
We find that challenging a stereotype by applying it broadly (e.g., to ``lots of people''; \textsc{Lots} and \textsc{Alts}; Figure~\ref{fig:intro}) is generally the most preferred strategy. 
In contrast, statements containing direct counter-evidence (e.g., \textsc{Dir-Ind} and \textsc{Dir-Grp}; Figure~\ref{fig:intro}) are the least popular. 
Additionally, we observe that the most favored strategy varies depending on whether the stereotype is explicitly presented to annotators (e.g., providing the essentialist belief in Figure~\ref{fig:intro}) or only conveyed implicitly (e.g., only providing the first statement in Figure~\ref{fig:intro}). For example, direct counter-evidence is more popular when the stereotype is explicitly provided. Our results highlight the complexity of countering essentialist beliefs and the importance of further investigation at the intersection of NLP and psychology. 

\section{
Automatically Countering Essentialism
} \label{sec:method}
We operationalize our counterstatement generation by focusing on the expression of stereotypes through generics (\S\ref{ssec:link-generics-stereotypes}). 
Inspired by work in psychology and philosophy, we construct five types of counterstatements to a stereotype (\S\ref{sec:countertypes}).

\begin{table*}[t]
    \centering
    \scalebox{0.85}{
    \begin{tabular}{ll}
        \hline
        \textbf{TEXT:} \textit{RT @Vbomb20: Got these hoes on my dick like brad pitt} & \textbf{GENERIC:} Women are sex objects.\\
        \hdashline
        \multicolumn{2}{l}{
            \begin{tabular}[t]{@{}l@{}}Actually this is a generalization about women. +\\
            \quad\quad \textbf{(\ref{template1}-\textsc{Grp})}\ The following women are not sex objects: businesswomen, female atheletes, and female movie stars.\\
            \quad\quad \textbf{(\ref{template1}-\textsc{Ind})}\ The following women are not sex objects: ellen degeneres, sarah palin, and rachel maddow.\\
            \quad\quad \textbf{(\ref{template2})}\ Men can also be sex objects.\\
            \quad\quad \textbf{(\ref{template3})}\ Lots of people can be sex objects.\\
            \quad\quad \textbf{(\textsc{Tol})}\ All groups of people deserve tolerance.\end{tabular}
        }\\
        \hline
        \textbf{TEXT:} \textit{What's black and doesn't work? Half of London} & \textbf{GENERIC:}  Black people don't work\\
        \hdashline
        \multicolumn{2}{l}{
        \begin{tabular}[t]{@{}l@{}}Actually, this is a generalization about black people. + \\ \quad\quad \textbf{(\ref{template1}-\textsc{Grp})}\ The following black people work: black businessmen, famous black people, and black movie stars.\\
        \quad\quad \textbf{(\ref{template1}-\textsc{Ind})}\ The following black people work: barack obama, misty copeland, and usain bolt.\\
        \quad\quad \textbf{(\ref{template2})}\ White folks may also not work.\\
        \quad\quad \textbf{(\ref{template3})}\ Lots of people don't work.\\
        \quad\quad \textbf{(\textsc{Tol})}\ All groups of people deserve tolerance. \end{tabular}
        }
        \\
        \hline
        \hline
        \begin{tabular}[t]{@{}l@{}}\textbf{TEXT:} \textit{How do you kill a thousand flies in one hit?}\\ \textit{Slap an Ethiopian in the face.}\end{tabular} & \textbf{GENERIC:}  Ethiopian people are dirty.\\
        \hdashline
        \multicolumn{2}{l}{
        \begin{tabular}[t]{@{}l@{}}Actually, this is a generalization about ethiopian people. + \\ \quad\quad \textbf{(\ref{template1}-\textsc{Grp})}\ The following ethiopian people are not dirty: male atheletes, female movie stars, \\\quad\quad\quad\quad\quad\quad\quad and people who practice judaism.\\
        \quad\quad \textbf{(\ref{template1}-\textsc{Ind})}\  The following ethiopian people are not dirty: kenyan marathon runners, michael jackson,\\\quad\quad\quad\quad\quad\quad\quad and ryan reynolds.\\
        \end{tabular}
        }
        \\
        \hline
        \begin{tabular}[t]{@{}l@{}}\textbf{TEXT:} \textit{A muslim enters a building..}\\ \textit{With 500 passengers and a plane}\end{tabular} & \textbf{GENERIC:}  Muslims are terrorists.\\
        \hdashline
        \multicolumn{2}{l}{
        \begin{tabular}[t]{@{}l@{}}Actually, this is a generalization about muslims. + \\ \quad\quad \textbf{(\ref{template1}-\textsc{Grp})}\ The following muslims are not terrorists: male muslim businessmen, muslims businessmen, \\\quad\quad\quad\quad\quad\quad\quad  and male muslim movie stars.\\
        \quad\quad \textbf{(\ref{template1}-\textsc{Ind})}\  The following muslims are not terrorists: adult muslim men, all muslims, and malala yousafzai.\\
        \quad\quad\quad\quad\quad\quad\quad $\bm{\hdots}$
        \end{tabular}
        }
        \\
        \hline
    \end{tabular}
    }
    \caption{Automatically generated counterstatements (\S\ref{sec:countertypes}) from our system. The bottom two examples illustrate challenges with factuality in the \textsc{Dir} counterstatements.}
    \label{tab:statementexs}
\end{table*}
\subsection{Stereotypes as Generics}\label{ssec:link-generics-stereotypes}
Many negative stereotypes are expressed as generics; they generalize a dangerous or harmful quality (e.g., being a drunkard) to an entire group (e.g., Scots) based on the behavior of only a few individuals. ~\citet{leslie2008generics,leslie2017original} termed such generics \textbf{striking} and argued that such generalizations are based upon an assumption that all members of the group in question (e.g., Scots) are \textit{disposed} to possess the dangerous or harmful quality. We argue that many stereotypes can also be interpreted as asserting a \textbf{quasi-unique} association between the group and quality. For example, ``Scots are drunkards''  also implies that Scots are distinctly more likely than other groups (e.g., the English) to exhibit drunkenness.
In our work, we assume that all stereotypes under consideration are generics and have both interpretations. 

Since generics are unquantified, they naturally allow for \textbf{exceptions} (i.e., counterexamples to the generic).
While these exceptions may provide a relevant source of counter-statements for a stereotype, some evidence from psychology suggests that people are adept at maintaining their stereotyped beliefs in the face of such specific exceptions \cite[e.g.,][]{kunda1995maintaining}. Therefore, we experiment with a variety of different counter-statements.

\subsection{Generating Counter-Speech}
\label{sec:countertypes}
To generate counter-speech to stereotypes, we produce five types of outputs in three broad categories (see Table~\ref{tab:statementexs}). Since the stereotypes we consider are expressed as generics (e.g., ``Scots are drunkards''), they can be separated into three components: a \textit{group} (e.g., Scots), a \textit{relation} (e.g., are), and a \textit{quality} (e.g., ``drunkards''), which we use to construct the counter-speech. Additionally, we prepend the sentence ``Actually, this is a generalization about \textsc{GROUP}'' to each type of statement we generate, in order to contextualize the statements as counter-speech.

\paragraph{Direct Exceptions (\textsc{Dir})}
Direct exceptions present subgroups or individuals that do not have the quality specified in the generic, and thereby counter the striking or extrapolating implications of the stereotype. For example, for ``Scots are drunkards'', the extrapolating implication is that ``\textit{All} Scots are drunkards''; thus, direct exceptions would be either individual Scots (e.g., Ewan McGregor\footnote{\url{https://fherehab.com/learning/celebrities-who-dont-drink}}) or sub-groups of Scots (e.g., Scottish babies) who are not drunkards. We follow~\citet{allaway2022penguins} who propose that these exceptions can be constructed with the following template:
\setlength{\abovedisplayskip}{3pt}
\setlength{\belowdisplayskip}{3pt}
\begin{align}
    &\textsc{Group}(x) + \text{not } \textit{relation} + \textsc{Quality}. \label{template1}\tag{\textsc{Dir}}
\end{align}
We say that $\textsc{Group}(x)$ is satisfied if $x$ is either a specific member of the group or a subgroup. 
We generate subtypes (i.e., subgroups and specific group members) using GPT-3~\citep{brown2020language}. In particular, we prompt GPT-3 with a list of subtypes for an example group not in our data and query the model to produce subtypes for \textsc{Group} as the prompts completion. We choose as our example group ``men'' (see Appendix~\ref{appsec:gpt3} for prompts).
We then construct exceptions following template~\ref{template1} using each generated subtype. 
In order to select the most truthful and relevant subtypes, we apply a truth discriminator from \citet{allaway2022penguins} 
to each exception, and rank the subtypes by the probability of being true and relevant. 
We construct the final statements by combining the top three ranked subgroups into a single exception ((\ref{template1}-\textsc{Grp}) in Table~\ref{tab:statementexs}) and combining the top three individuals into a single exception ((\ref{template1}-\textsc{Ind}) in Table~\ref{tab:statementexs}).

\paragraph{Broadening Exceptions (\textsc{Alts})}
Broadening exceptions challenge the quasi-unique implication of the generic by attributing the quality in question to a different social group (e.g., ``Americans can also be drunkards''). ~\citet{allaway2022penguins} propose that these exceptions follow the template:
\begin{align}
    &{\nsim}\textsc{Group}(x) + \textit{relation} + \textsc{Quality}.\label{template2}\tag{\textsc{Alt}}
\end{align}
where ${\nsim}\textsc{Group}$ indicates a contextually relevant alternative group. For example, if \textsc{Group} = \textsc{Scots}, then a contextually relevant alternative would be ${\nsim}\textsc{Group}$ = \textsc{Americans}. In our work, we define the relevant alternative group ${\nsim}\textsc{Group}$ to be the perceived oppressing group. For example, if the generic is ``women are vain'', then ``men'' would be the relevant alternative group ${\nsim}\textsc{Women}$ (i.e., the oppressing group). To avoid generating stereotypes about the oppressing group, we convert the relation into a hedged form (see Appendix~\ref{appsec:dataprocess}). For example, if the relation is ``are'', the hedged form of the relation would be ``can be''. 

\begin{figure*}[t!]
    \centering
    \begin{subfigure}[b]{0.5\textwidth}
        \centering
        \includegraphics[width=\textwidth]{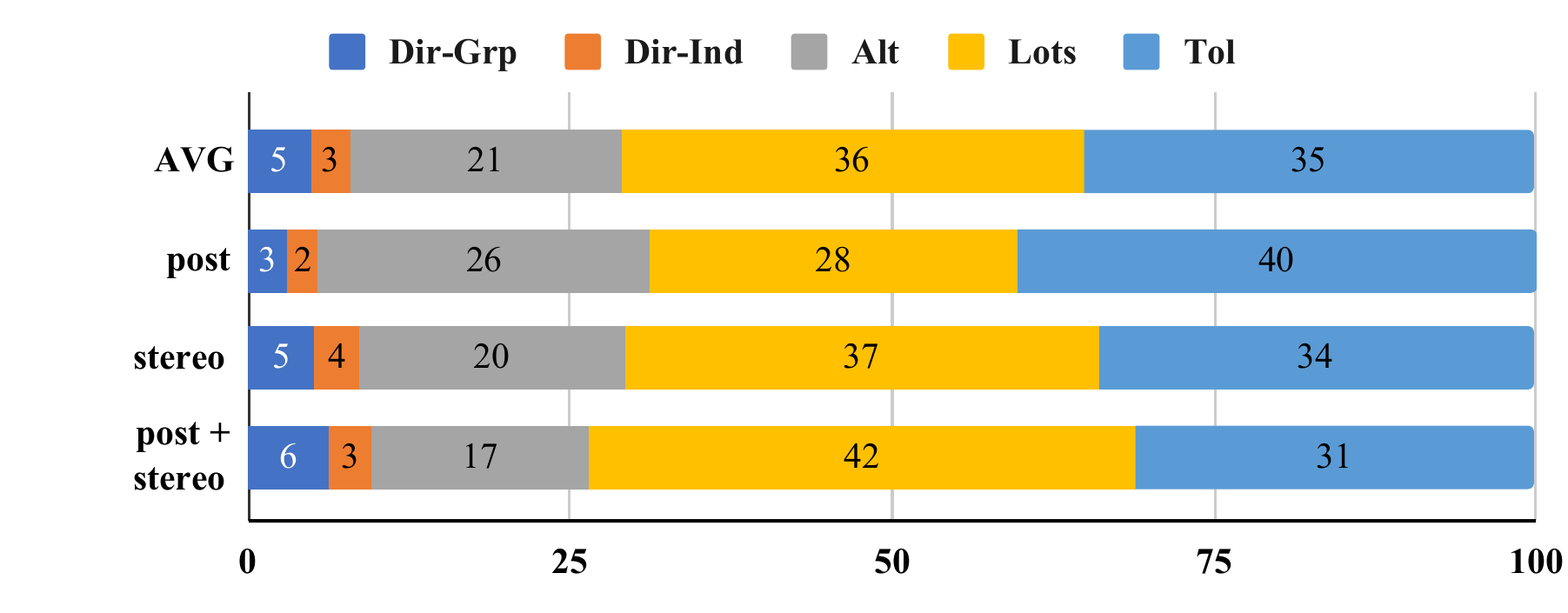}
        \caption{First choice.}
        \label{subfig:firstchoice}
    \end{subfigure}%
    \begin{subfigure}[b]{0.5\textwidth}
        \centering
        \includegraphics[width=\textwidth]{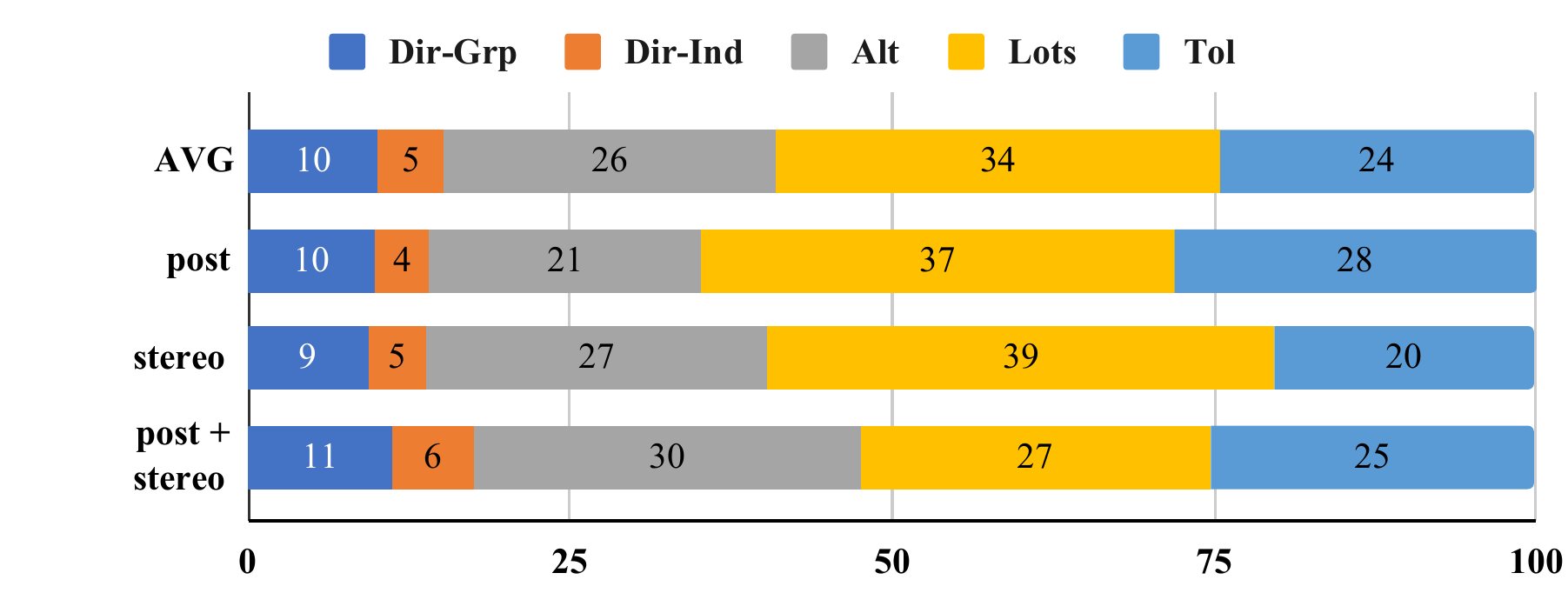}
        \caption{Second choice.}
        \label{subfig:secondchoice}
    \end{subfigure}
    \caption{Percentage of annotators that selected each counterstatement type (\S\ref{sec:method}) across all three settings.
    }
    \label{fig:choicecharts}
\end{figure*}

\paragraph{Broadening Universals (\textsc{Lots})}
In addition to broadening exceptions, we generate \textit{broadening universals}, which maximize the scope of the quality so that it includes people in general, rather than any specific social group. That is, we generate statements following:
\begin{align}
    \text{Lots of people} + \textit{relation} + \textsc{Quality}.\label{template3}\tag{\textsc{Lots}}
\end{align}
For example, ``Lots of people are drunkards" is a broadening universal for the stereotype ``Scots are drunkards''. See (\ref{template3}) in Table~\ref{tab:statementexs}. Similarly to the statements following template~\ref{template2}, we also hedge the relation in template~\ref{template3}.

\paragraph{Tolerance (\textsc{Tol})}
Finally, we include the denouncing statement, ``All groups of people deserve tolerance'', since denouncing is a common strategy in countering hate-speech~\cite[e.g.,][]{mathew2019thou,qian2019benchmark,Ziegele2018JournalisticCI}. This form of counter-speech does not depend on the details of the generic in question and so is the same for all stereotypes. See (\textsc{Tol}) in Table~\ref{tab:statementexs}.

\section{Online Study}\label{sec:experiments}
As a preliminary investigation into the task of generating counterstatements to combat essentialism, we use posts with gold-annotated implications (\S\ref{ssec:sbf-data}) to conduct an online experiment with crowdworkers (\S\ref{ssec:mturk-study-setup}).

\subsection{Essentialism Data}\label{ssec:sbf-data}
We use annotations provided in the SBIC~\citep{sap2020socialbiasframes} to obtain pairs $(t, s)$ where $t$ is a text and $s$ is a stereotype implied by $t$ (i.e., an essentialist implication that can be drawn from $t$). The $s$ in SBF are human written and so to ensure the statements we consider are clear implications of the text $t$,
we use only instances where at least two out of the three human annotators wrote the same stereotype verbatim. This results in a set of $227$ pairs, covering $25$ unique groups, where each $s_i$ can be clearly inferred from $t_i$. 

\subsection{Study Setup}\label{ssec:mturk-study-setup}
In order to investigate the effectiveness of different counter statements (\S\ref{sec:method}), we conduct three different human studies. In each study, we ask annotators on Amazon Mechanical Turk to play the role of an online content moderator or fact-checker whose job is to provide counterstatements to expressed stereotypes. Each annotator is provided with a statement and a set of machine-generated counterstatements and asked to select their first and second choices. We also include an attention check to monitor annotation quality, and collect information on how much annotators agree with the provided statement and annotator demographic information. See full instructions in Appendix~\ref{appsec:annotation}.

Our three human studies vary the statements provided to annotators: \textbf{(1) post} -- an original text $t$ from SBF, \textbf{(2) stereo} -- the stereotype $s$ implied by a text $t$, or \textbf{(3) post + stereo} -- both $t$ and $s$. Note that for each pair $(t, s)$ the counterstatements are always derived from $s$, regardless of whether annotators are provided $s$ directly.

\section{Empirical Results}\label{ssec:results}
Our results show clear differences in how often certain types of counterstatements are preferred over others to combat essentialism (Figure~\ref{fig:choicecharts}).
We see that overall, the \textsc{Lots} counterstatements are the most popular for both first and second choice. 
In addition, when considering broadening statements grouped together (\textsc{Lots} and \textsc{Alt}), there is a clear preference for such statements, compared to both the \textsc{Tol} and the direct exceptions. 
Despite the lack of content in the \textsc{Tol} statements, these are the second most popular as the first choice. 
Note, we choose not to conduct statistical tests because our goal is not to find the single most effective countering strategy but rather to study a range of strategies.

\begin{figure}
    \centering
    \includegraphics[width=\columnwidth]{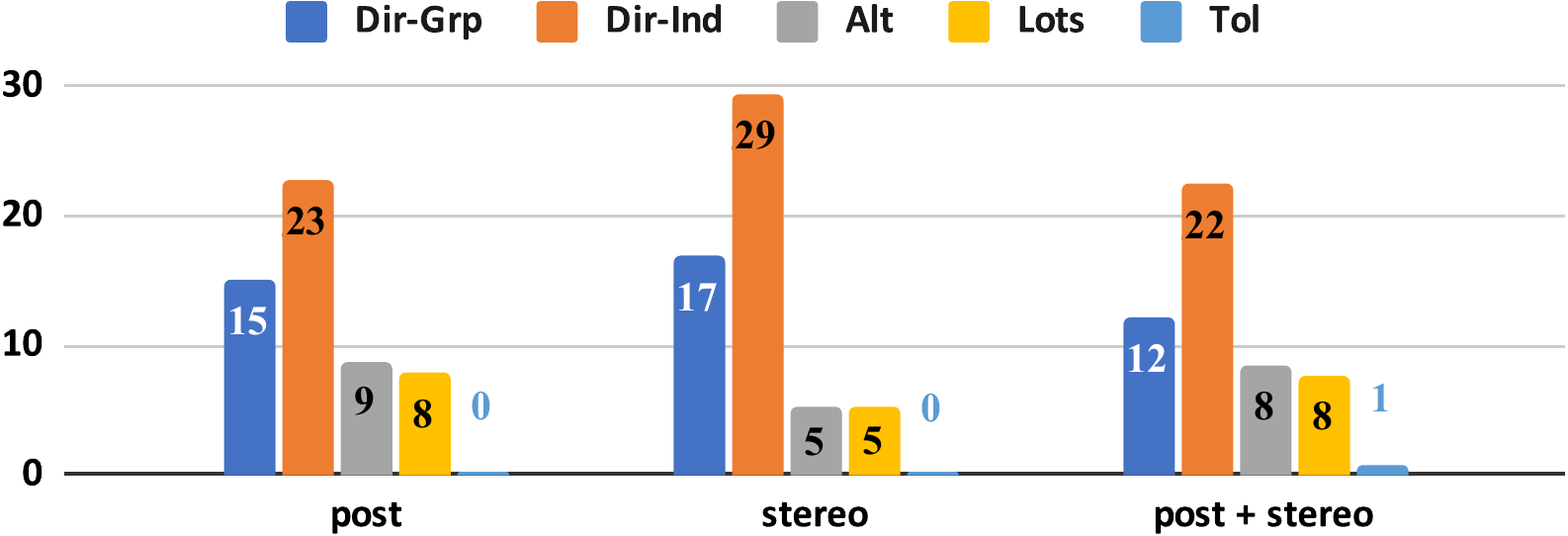}
    \caption{Percentage of counterstatements marked as incorrect for each setting. Counterstatements are the same across settings, variation is due to annotators.}
    \label{fig:percentincorrect}
\end{figure}

Of the generics-exceptions-based counterstatements, the direct exceptions \textsc{Dir} are consistently the least preferred. We hypothesize that this is impacted by the high portion of incorrect statements among the \textsc{Dir} type (Figure~\ref{fig:percentincorrect}), as well as the subjective nature of many stereotypes (e.g., in Table~\ref{tab:statementexs}, being a `sex object' is subjective). 
When considering only the statements \textit{not} marked as incorrect by annotators, we do not observe a change in relative popularity. Therefore, future investigation is needed to understand the role of correct individuals in counterstatements.

In contrast, the broadening exceptions \textsc{Alt} rank second as the second-choice and only 7\% are marked as incorrect. 
We also note that in settings where the stereotype is provided explicitly (stereo and stereo+post) the proportion of \textsc{Lots} was higher (and \textsc{Tol} lower) for the first choice, and for the second choice the proportion of \textsc{Alt} increased markedly. 
From this we observe that the effectiveness of a countering strategy may depend on the explicitness of the demonstrated bias. 
For example, generalizing the stereotype (\textsc{Lots}) may be less effective when the stereotype is not explicitly identified (post setting).

\begin{figure}
    \centering
    \begin{subfigure}[b]{\columnwidth}
        \centering
        \includegraphics[width=\textwidth]{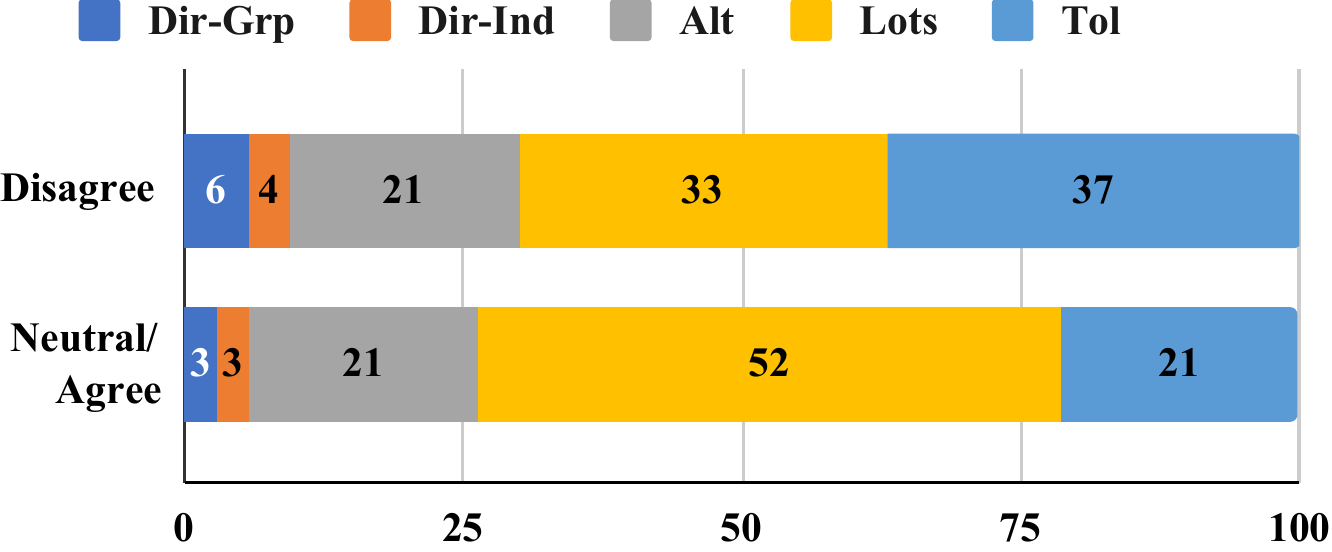}
        \caption{Percent of annotators that selected each counterstatement type as their first choice.}
        \label{subfig:agreestatements}
    \end{subfigure}
    \begin{subfigure}[b]{\columnwidth}
        \centering
        \includegraphics[width=\textwidth]{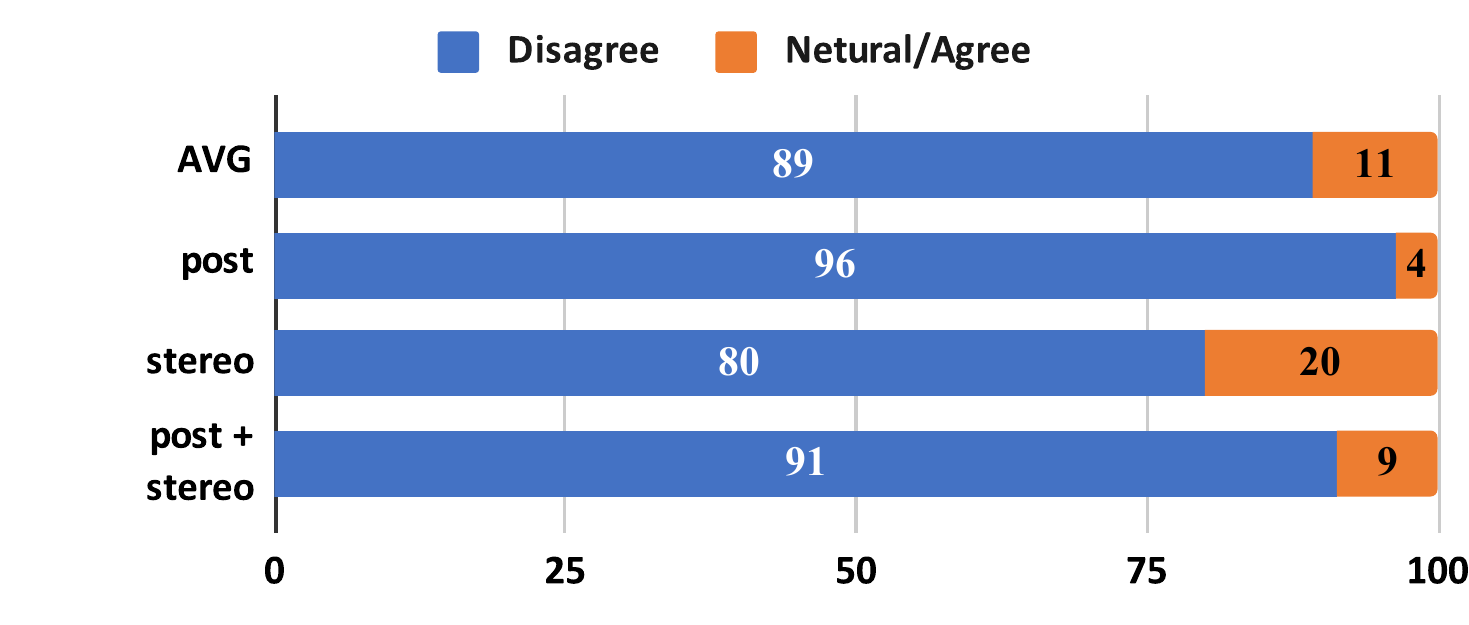}
        \caption{Percent of annotators who agreed with the statement.}
        \label{subfig:annotatoragree}
    \end{subfigure}
    \caption{Self-reported annotator agreement with the provided statement(s).}
    \label{fig:agree}
\end{figure}

Finally, we observe that when annotators agree with a statement, their preference for \textsc{Lots} statements increases while the preference for \textsc{Dir} counterstatements decreases (Fig.~\ref{subfig:agreestatements}). Annotator preference for \textsc{Tol} also decreases. 
We also note that annotators more often endorse a belief when it is stated explicitly, rather than implied by a text (Fig.~\ref{subfig:annotatoragree})
These results underscore the importance both of directly identifying an essentialist belief from an implication and of reasoning about the implications of the stereotype when countering real-world essentialist beliefs (i.e., from individuals who endorse the belief).

\section{Discussion and Conclusion}\label{sec:discussion}
Through our online studies, we find that broadening statements are the most preferred type of counterstatement, while statements with direct counter-evidence are consistently least preferred. In addition, we observe variation across our three settings. Below, we discuss how are findings related to work in psychology (\S\ref{ssec:psych}) and content moderation (\S\ref{ssec:hate-speech}), and finally, outline challenges, limitations, and future directions (\S\ref{ssec:future}).

\subsection{Stereotypes and Psychology}\label{ssec:psych}
Generic language, with its quasi-unique implications, readily conveys essentialist beliefs. 
Indeed, psychological research shows that generic language is a powerful mechanism by which social essentialist beliefs are \textit{transmitted} between people, and even across generations \cite{rhodes2012cultural,leshin2021does}.
Such implications can have a profound impact on children  --- e.g., girls as young as 6 years old have absorbed the stereotype that males are more likely than females to be ``really, really smart'' \cite{bian2017gender}. 
In order to challenge such essentialist beliefs, we argue that it is important to consider the complexities of generics and associated inferences. 

Through reasoning directly about the implications of generics, we construct counterstatements that directly challenge essentialist implications. In particular,
our results highlight the value of broadening statements (\textsc{Lots} and \textsc{Alt}), which counter
the implication that a particular negative quality
is distinctive of a particular group (e.g., ``\textit{Only} women are vain''). This finding is consistent with recent work in psychology, in particular \cite{foster-hanson_leslie_rhodes_2019}.
These statements thereby challenge the cognitive \textit{value} of the stereotype as an information-processing short-cut~\cite{Devine1989StereotypesAP}, since the wide applicability of the stereotyped quality may result in many incorrect inferences (e.g., assuming someone is not vain because they are not a woman). 

Furthermore, our results corroborate findings from psychology that individuals who do not fit a stereotype are not viewed as invalidating that stereotype, since they are categorized as special~\cite[e.g.,][]{kunda1995maintaining}. In particular,  the consistently low preference for direct exception statements comports with that finding (\textsc{Dir-Ind} and \textsc{Dir-Grp}). Although providing facts (e.g., exceptional individuals) has been previously studied as a strategy to counter hate-speech~\cite[e.g.,][]{chung-etal-2019-conan,mathew2019thou}, our work specifically isolates the \textit{type} of facts (i.e., direct counter-evidence versus broadening statements) as a variable for investigation. 
As such, we can observe that providing broadening facts is much more effective than counter-evidence. This further highlights the importance of reasoning about the specific implications of a text to counter essentialist beliefs.

\subsection{Essentialism, Counter Hate-Speech, and Content Moderation}\label{ssec:hate-speech}
Although countering essentialism is similar in spirit to countering hate-speech and content moderation, common strategies in the latter are often inapplicable to countering essentialist beliefs.
In content moderation, discursive actions such as answering clarifying questions or providing additional details are common~\cite{Ziegele2018JournalisticCI}.
However, since essentialist beliefs are often conveyed implicitly (e.g., see statements in Figure~\ref{fig:intro}), discursive actions aimed at a text may not actually address its essentialist implications. For example, the additional detail \textit{``libt$^*$rd is not a real language''} does not actually counter the implication that \textit{liberals are stupid} in Fig~\ref{fig:intro}. Similarly, while humor, expressing affiliation with the targeted group (e.g., \textit{``us Scots only having a wee cuppa tea''}), and pointing out hypocrisy or contradictions (e.g., \textit{``it needs to involve food to be a meal''}) are common when countering hate-speech~\cite{chung-etal-2019-conan,mathew2019thou}, they also do not address the essentialist beliefs implicit in a text (e.g., that \textit{Scots are drunkards}, Figure~\ref{fig:intro}).
As such, we argue that it is important to investigate effective ways to counter essentialist implications, as distinct from general counter-speech and content moderation.

\begin{figure}
    \centering
    \begin{subfigure}[b]{\columnwidth}
        \includegraphics[width=\textwidth]{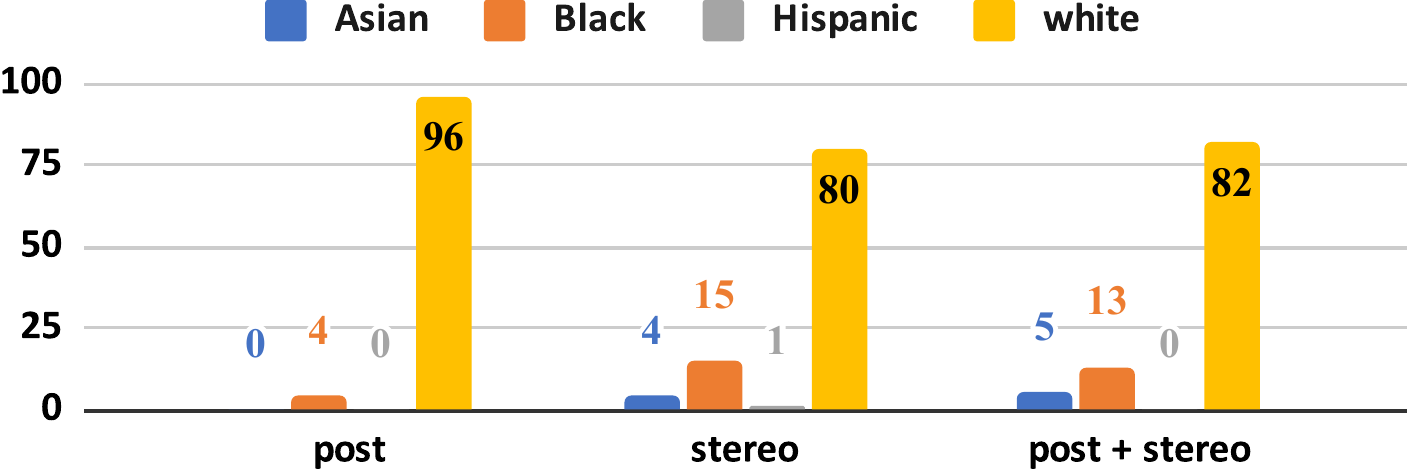}
        \caption{Racial demographics.}
        \label{subfig:race}
    \end{subfigure}
    \begin{subfigure}[b]{\columnwidth}
        \includegraphics[width=\textwidth]{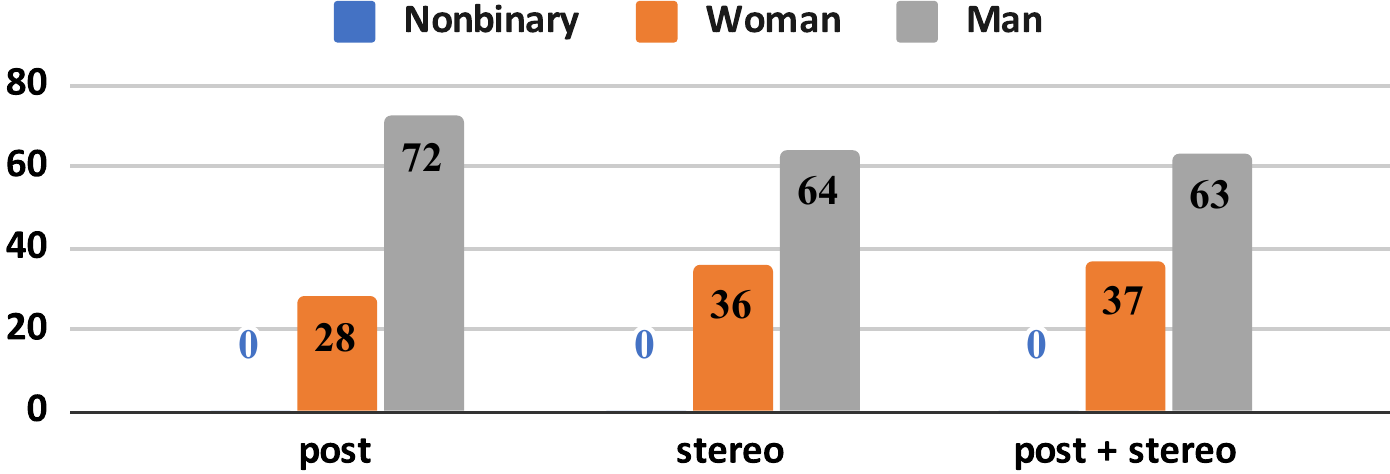}
        \caption{Gender demographics.}
        \label{subfig:gender}
    \end{subfigure}
    \caption{Self-reported annotator demographics (percentage) across settings.
    }
    \label{fig:annotatordemos}
\end{figure}
\subsection{Limitations, Challenges, and Future Directions}\label{ssec:future}

Along with promising preliminary findings, our results highlighted several limitations and challenges that should be tackled in future work.

\paragraph{Human-annotated implications}
Since this work constitutes preliminary investigation on the promise of using NLP tools for combating essentialism, we used a corpus of statements paired with gold human-annotated implications. 
However, such annotations will not always be available. 
Future work should examine whether our findings would hold with machine-generated implications \cite[e.g., using the neural model from][]{sap2020socialbiasframes}, on various types of source domains and overtness levels \cite[e.g., the corpus of implicit toxicity from][]{hartvigsen2022toxigen}.
Furthermore, future research could investigate how the quality and specificity of the implications affects the counterstatement generation and effectiveness.

\begin{table}[t]
    \centering
    \scalebox{0.8}{
    \begin{tabular}{l|r}
        \hline
        \textbf{Group} & \textbf{Nb Examples} \\
        \hline
        Black folks & 66\\
        Women & 60\\
        \hdashline
        Muslim folks & 18\\
        Jewish folks & 16\\
        Asian folks & 15\\
        \hdashline
        Gay men & 7\\
        Latino/Latina folks & 6\\
        Liberals & 5\\
        Feminists & 4\\
        \hdashline
        African folks & 3\\
        Mentally disabled folks & 3\\
        Indian folks & 3\\
        Lesbian women & 3\\
        Immigrants & 3\\
        Ethiopian folks & 3\\
        American folks & 2\\
        Mexican folks & 2\\
        \hdashline
        Physically disabled folks & 1\\
        Folks with mental illness/disorder & 1\\
        Japanese folks & 1\\
        Polish folks & 1\\
        Arabic folks & 1\\
        Italian folks & 1\\
        Christian folks & 1\\
        Native American/First Nation folks & 1\\
        \hline
    \end{tabular}
    }
    \caption{Counts for number of examples per group. There are 227 examples total across 25 unique groups.}
    \label{tab:groupstats}
\end{table}

\paragraph{Targeted group and annotator identity}
Our studies are conducted on Amazon Mechanical Turk which can be lacking in diversity among annotators. For example, the majority of annotators in our study were white (Fig~\ref{subfig:race}) or male (Fig~\ref{subfig:gender}). In contrast, targeted groups are often \textit{not} white or male (see Table~\ref{tab:groupstats}). Since an annotator's identity and beliefs may impact their perceptions of how effective a counterstatement is~\cite[as they do with perceptions of toxicity;][]{sap2022annotatorsWithAttitudes}, homogeneity in the annotator population limits our results. Additionally, how deeply rooted an essentialist belief is for an annotator may impact what they consider effective counterstatements. 
Our results, which show large variation in annotator preference depending on whether they endorse a statement, corroborate these findings.
Therefore, future work should investigate more diverse annotator pools or matching annotators to targeted groups, as well as examining how annotator's familiarity with essentialist beliefs and identities affect their judgements.

Furthermore, prior work in countering hate-speech has show that effective strategies can vary widely depending on the target group~\citep{mathew2019thou,chung-etal-2019-conan}. 
In our work, we consider results aggregated across all groups.
However, community-specific investigations are an important future step towards developing effective counter-statements. 

\paragraph{Accuracy of generated exceptions}
The selection of specific individuals for direct exceptions presents an ongoing challenge, based on the high number of \textsc{Dir-Ind} marked incorrect. 
Since language models often encode biases and stereotypes derived from training corpora \cite{sheng2019woman}, they may have difficulty producing \textit{relevant} individuals who are not prototypical (i.e., they do not have a particular stereotype).
We illustrate incorrect individuals and subgroups in the bottom two examples of Table~\ref{tab:statementexs}. 
Additionally, as mentioned in \S\ref{ssec:results}, many stereotypes are subjective (e.g., ``women are vain''). 
Therefore, individuals who are counterexamples to the stereotype may be judged differently by different people (e.g., our system proposes that ``taylor swift, sarah palin, and scarlett johansson'' are not vain).
Producing accurate and relevant direct exceptions to a stereotype is important for understanding the role of such examples to counter essentialist beliefs.

Our results and discussion highlight the complexity of countering essentialist beliefs.
We propose that future work should improve the factuality of counterstatements, particularly of direct counter-evidence, and consider both variation in respondent demographics and community-specific needs. Therefore, we argue that working at the intersection of NLP and psychology is crucial for further investigations in this area.

\section{Societal and Ethical Considerations}\label{sec:ethical-considerations}
\paragraph{Annotation Considerations}
Prior work has highlighted the potential harms to workers who are subjected to offensive statements~\cite{Roberts2017-rp,Steiger2021-ka}. To mitigate these, we encourage annotators to reach out to the authors with concerns and questions or to the Crisis Text Line.\footnote{\url{https://www.crisistextline.org/}} Additionally, our study design was approved by our ethics review board (IRB) and 
workers earned a median wage of \$10/h.

\paragraph{Risks of Generation}
Since our system automatically generates counterstatements, there is potential for misuse in several ways. First, our system can automatically and quickly produce millions of counterstatements could therefore be used in a distributed-denial-of-service attack. Second, by generating counterstatements to stereotypes in text the original text remains available and so it may still cause harm~\cite{Ullmann2019quarantining} and perpetuate essentialist beliefs. Additionally, the automatic construction of counterstatements has the potential to produce false statements and further harmful generalizations (e.g., generalize a harmful stereotype to another marginalized group). Considering these factors, it is important to jointly develop regulation alongside AI technology to limit harms and misuse in deployment~\cite{Crawford2021-kz,Reich2021-xw}.

\section*{Acknowledgements}
We would like to thank the Beaker Team
at AI2 for  the compute infrastructure, and the anonymous reviewers for their suggestions. This work is supported in part by the National Science Foundation Graduate Research Fellowship under Grant No. DGE-1644869. The views and conclusions contained herein are those of the authors and should not be interpreted as necessarily representing the official policies, either expressed or implied, of the NSF or the U.S. Government. The U.S. Government is authorized to reproduce and distribute reprints for governmental purposes notwithstanding any copyright annotation therein.

\bibliography{00-refs}

\appendix 
\section{Data Processing}\label{appsec:dataprocess}
To construct the hedged counterstatments, if the main verb is `is' or `are' we convert it to `can also be'. For example `men are vain' becomes `men can also be vain'. If the main verb is `should' we convert it to `should also'. Otherwise, we insert `may also' before the quality. For example, `men think they know everything' becomes `men may also think they know everything'.

We also note that the group names in Table~\ref{tab:groupstats} have been normalized. We will include both the normalized and unnormalized names in the released data.

\subsection{GPT-3 Generation}\label{appsec:gpt3}
We access GPT-3 using the API from OpenAI\footnote{\url{https://beta.openai.com/docs/introduction}}. To obtain subtypes from GPT-3 we use the \textit{`davinci'}
model and top-$p$ sampling with $p = 0.9$, temperature 0.8 and maximum length $100$ tokens. The presence and frequency penalties are both $0$. We kept the top 5 generations from GPT-3. 
We filter out generations that are the same as the queried group. The prompts are shown in Table~\ref{tab:gpt3prompts}. We randomized the order of the 5 examples in each prompt for every group. 

\begin{table*}[t]
    \centering
    \begin{tabular}{l|l}
        \textbf{Counterstatement Type} & \textbf{Prompt} \\ \hline
        \textsc{Dir-Grp} & 
        \begin{tabular}[t]{@{}l@{}} Consider the following groups of men:\\1. male students\\2. male authors\\3. male atheletes\\4. businessmen\\5. male movie stars\\\#\#\\\#\#\\Consider the following groups of \textsc{GROUP}:\\\end{tabular}\\
        \hline
        \textsc{Dir-Ind} &
        \begin{tabular}[t]{@{}l@{}} Consider the following groups of men:\\1. Barack Obama\\2. Sherlock Holmes\\3. Usain Bolt\\4. Ryan Reynolds\\5. Stephan Hawking\\\#\#\\\#\#\\Consider the following groups of \textsc{GROUP}:\\\end{tabular}\\
        \hline
    \end{tabular}
    \caption{Prompts for generating subtypes for \textsc{GROUP} from GPT3 (e.g., \textsc{GROUP}=women).}
    \label{tab:gpt3prompts}
\end{table*}

\section{Human Studies}\label{appsec:annotation}
For our user studies, we recruit annotators from Amazon Mechanical Turk who were qualified for a toxicity explanation task from our previous work \cite{anon}.\footnote{Anonymized to preserve double-blindness of reviewing, will be de-anonymized upon public release.}
Racial and gender breakdowns of our annotator pool are in Figure~\ref{fig:annotatordemos}.
Annotators were paid \$0.27 per task. 
For each instance in each of the three settings we have 3 annotators.
This study was approved by our institution's ethics board (IRB).

\begin{figure}
    \centering
    \fbox{\includegraphics[width=.95\columnwidth]{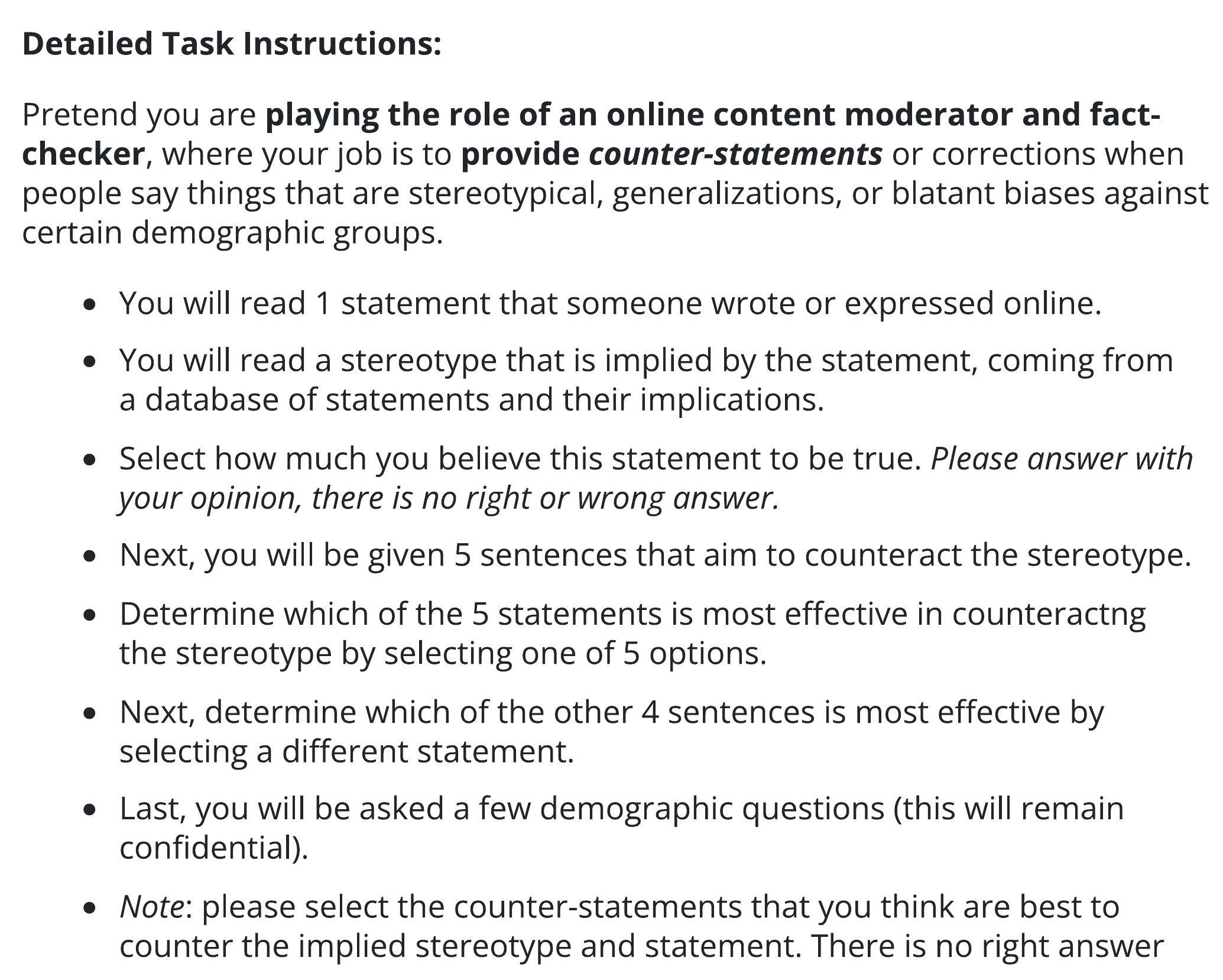}}
    \caption{Detailed annotation instructions for human studies.}
    \label{fig:annotationinstructions}
\end{figure}

\begin{figure}
    \centering
    \begin{subfigure}[b]{\columnwidth}
        \centering
        \fbox{\includegraphics[width=.95\textwidth]{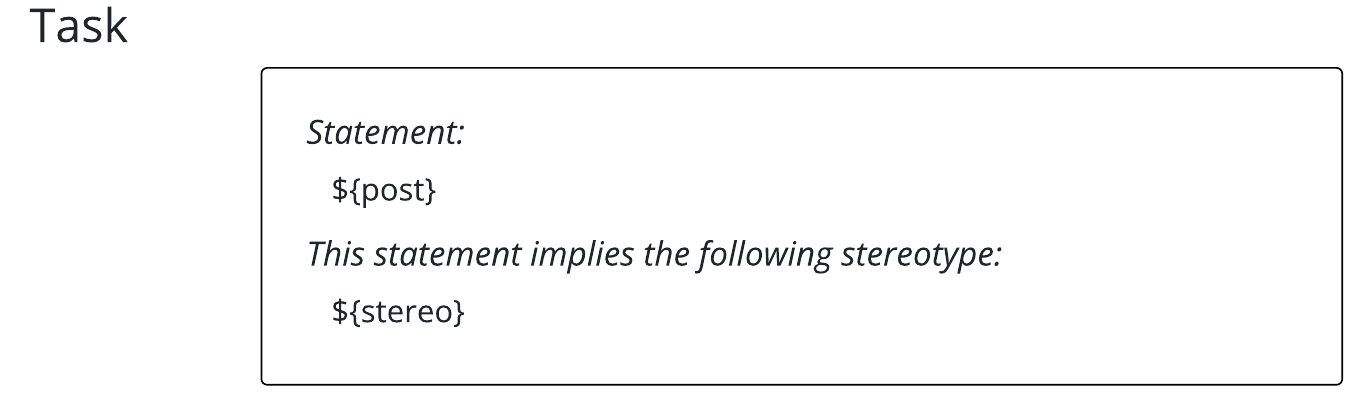}}
        \caption{Input presentation for \textit{post+stereo} setting. The statement was removed for the \textit{stereo} setting and the stereotype was removed in the \textit{post} setting.}
        \label{fig:task}
    \end{subfigure}
    \begin{subfigure}[b]{\columnwidth}
        \centering
        \fbox{\includegraphics[width=.95\textwidth]{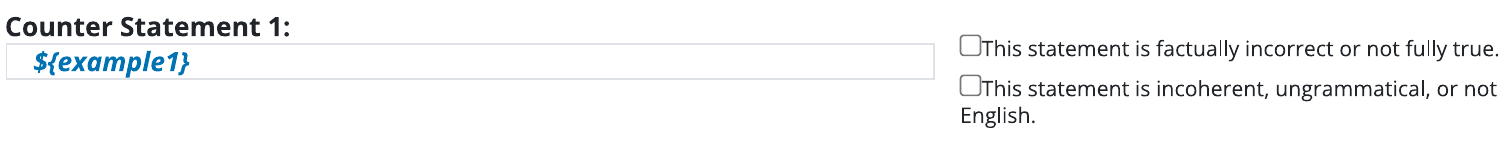}}
        \caption{Example presentation. All five types of counter statements are listed in the same manner.}
        \label{fig:taskexample}
    \end{subfigure}
    \begin{subfigure}[b]{\columnwidth}
        \fbox{\includegraphics[width=.95\textwidth]{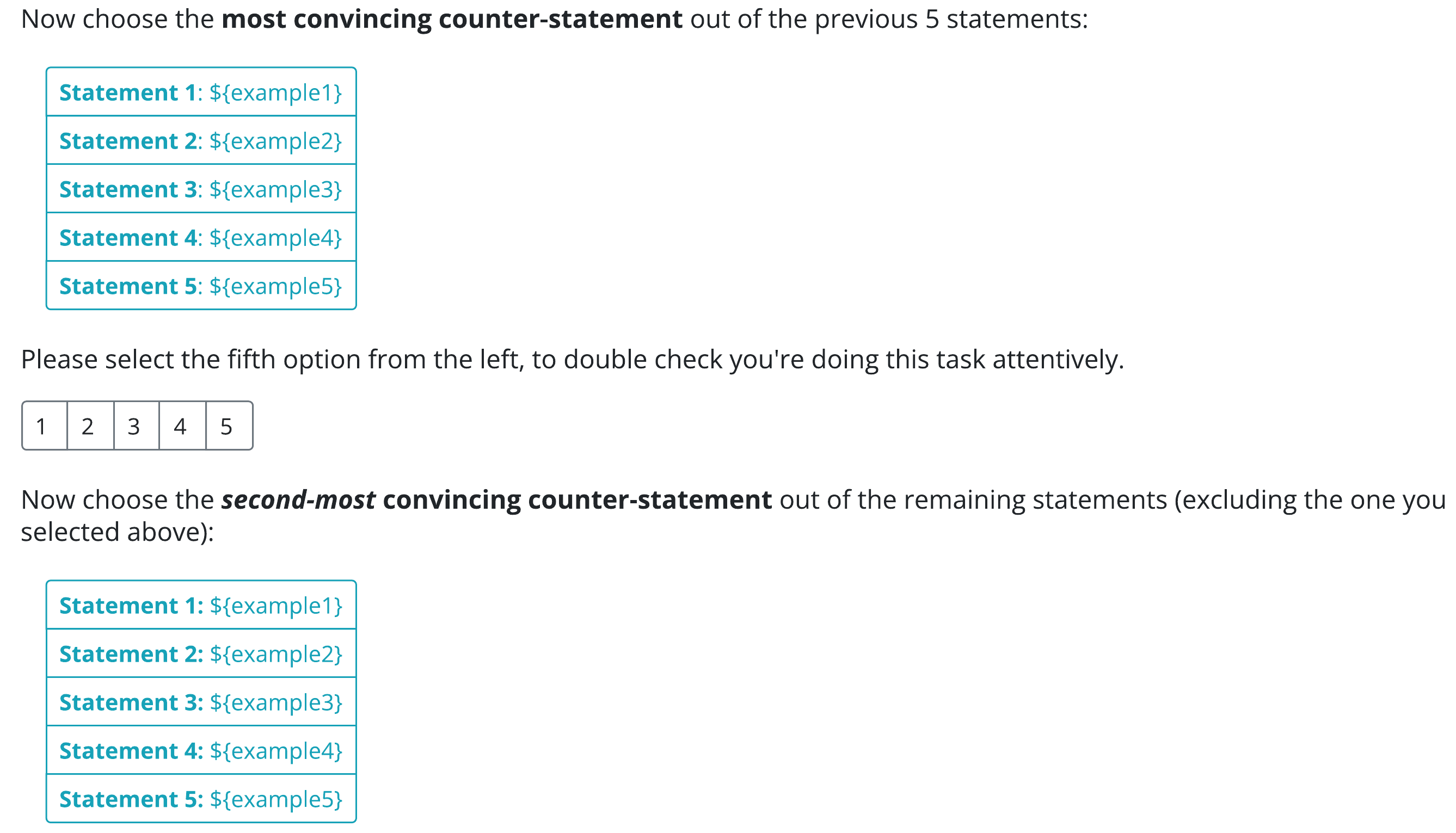}}
        \caption{Annotation questions and attention check.}
        \label{fig:taskquestions}
    \end{subfigure}
    \caption{Details of the annotation task for human studies.}
    \label{fig:mturk}
\end{figure}

We show the detailed task instructions in Figure~\ref{fig:annotationinstructions}. 
An example of the task setup is shown in Figure~\ref{fig:mturk}.
Before choosing the most convincing counter statements, annotators have the option to mark each statement as incorrect or ungrammatical (Figure~\ref{fig:taskexample}).
Note that before asking annotators to select their second choice, we include an attention check (in Figure~\ref{fig:taskquestions}). The attention check was randomly set in each HIT. Annotations where the attention check incorrect were discarded. As a result, we removed 3 annotations from the \textit{post} setting, 5 from the \textit{stereo} setting, and 4 from \textit{post+stereo}. 

\begin{figure}
    \centering
    \fbox{\includegraphics[width=.95\columnwidth]{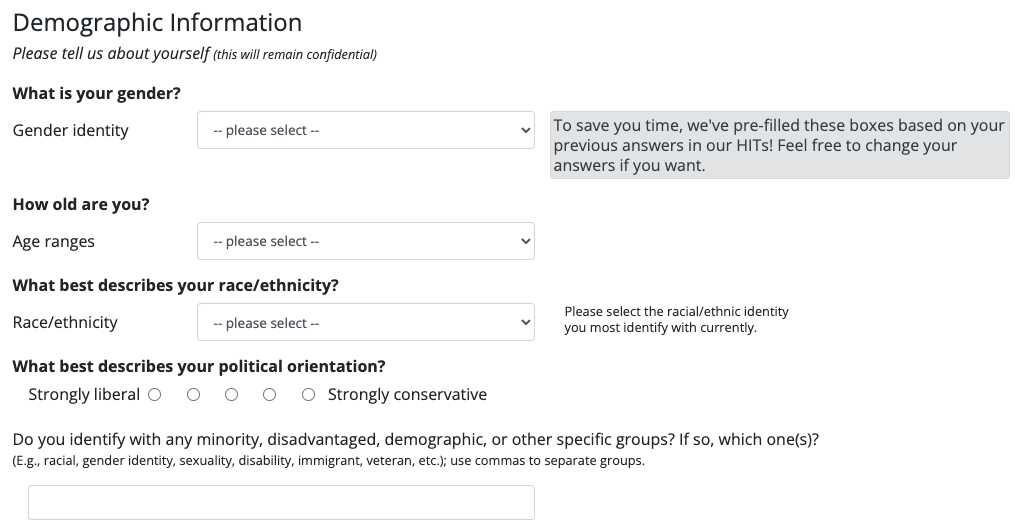}}
    \caption{Demographic questionnaire in human studies.}
    \label{fig:demoquestions}
\end{figure}

\begin{figure}
    \centering
    \fbox{\includegraphics[width=.95\columnwidth]{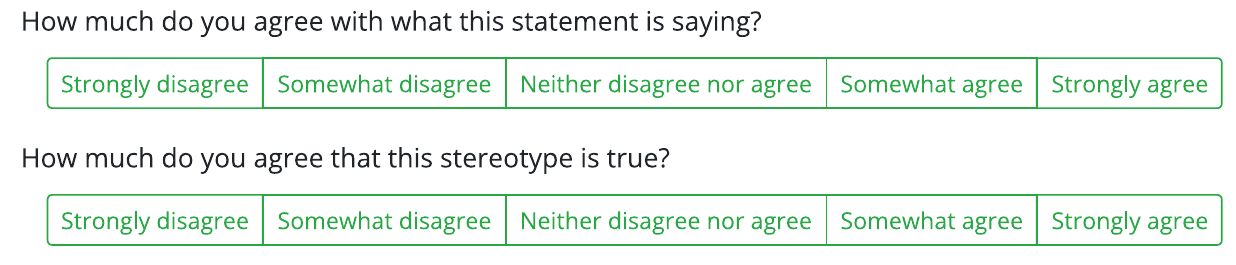}}
    \caption{Questions about stereotype belief of annotators.}
    \label{fig:agreequestions}
\end{figure}

For each annotation, we also collected demographic information (Figure~\ref{fig:demoquestions}). The demographic information is associated only with an annonymized annotator ID. Additionally, before annotators select counter-statements, we ask annotators to indicate their own belief in or agreement with the provided statement and stereotype (Figure~\ref{fig:agreequestions}).

\end{document}